# Facial Expression Classification Based on Multi Artificial Neural Network and Two Dimensional Principal Component Analysis

**Thai Le[1], Phat Tat[1] and Hai Tran[2]**

[1] Computer Science Department, University of Science, Ho Chi Minh City, Vietnam

[2] Informatics Technology Department, University of Pedagogy, Ho Chi Minh City, Vietnam

**Abstract**

Facial expression classification is a kind of image classification and it has received much attention, in recent years. There are many approaches to solve these problems with aiming to increase efficient classification. One of famous suggestions is described as first step, project image to different spaces; second step, in each of these spaces, images are classified into responsive class and the last step, combine the above classified results into the final result. The advantages of this approach are to reflect fulfill and multiform of image classified. In this paper, we use 2D-PCA and its variants to project the pattern or image into different spaces with different grouping strategies. Then we develop a model which combines many Neural Networks applied for the last step. This model evaluates the reliability of each space and gives the final classification conclusion. Our model links many Neural Networks together, so we call it Multi Artificial Neural Network (MANN). We apply our proposal model for 6 basic facial expressions on JAFFE database consisting 213 images posed by 10 Japanese female models.

## 1. Introduction

There are many approaches apply for image classification. At the moment, the popular solution for this problem: using K-NN and K-Mean with the different measures, Support Vector Machine (SVM) and Artificial Neural Network (ANN).

K-NN and K-Mean method is very suitable for classification problems, which have small pattern representation space. However, in large pattern representation space, the calculating cost is high.

SVM method applies for pattern classification even with large representation space. In this approach, we need to define the hyper-plane for classification pattern [1]. For example, if we need to classify the pattern into L classes, SVM methods will need to specify 1+ 2+ … + (L-1) = L (L-1) / 2 hyper-plane. Thus, the number of hyper-planes will rate with the number of classification classes. This leads to: the time to create the hyper-plane high in case there are several classes (costs calculation).

Besides, in the situation the patterns do not belong to any in the L given classes, SVM methods are not defined [2]. On the other hand, SVM will classify the pattern in a given class based on the calculation parameters. This is a wrong result classification.

One other approach is popular at present is to use Artificial Neural Network for the pattern classification. Artificial Neural Network will be trained with the patterns to find the weight collection for the classification process [3]. This approach overcomes the disadvantage of SVM of using suitable threshold in the classification for outside pattern. If the patterns do not belong any in L given classes, the Artificial Neural Network identify and report results to the outside given classes.

In this paper, we propose the Multi Artificial Neural Network (MANN) model to apply for image classification.

Firstly, images are projected to difference spaces by Two Dimensional Principal Component Analysis (2D-PCA).

Secondly, in each of these spaces, patterns are classified into responsive class using a Neural Network called Sub Neural Network (SNN) of MANN.

20

Lastly, we use MANN's global frame (GF) consisting some Component Neural Network (CNN) to compose the classified result of all SNN.

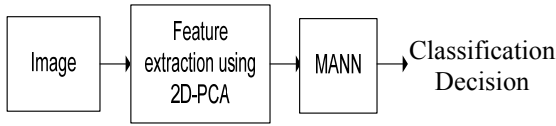

Fig 1. Our Proposal Approach for Image Classification

## 2. Background and Related Work

There are a lot of approaches to classify the image featured by m vectors $X = (v_1, v_2, ..., v_m)$. Each of patterns is needed to classify in one of L classes: $\Omega = \{\Omega i \mid 1 \leq i \leq L\}$. This is a general image classification problem [3] with parameters (m, L).

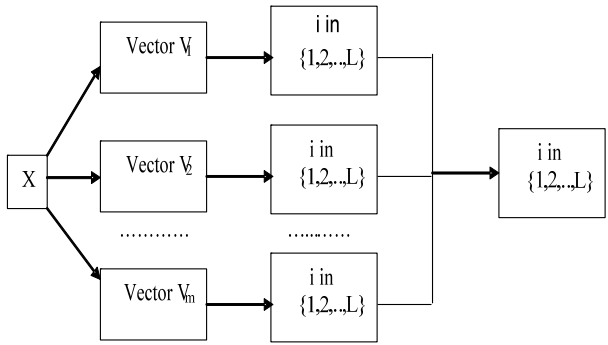

Fig 2. Image with m feature vectors Classification

First, the extraction stage featured in the image is performed. It could be used wavelet transform, or Principal Component Analysis (PCA). PCA known as one of the well-known approach for facial expression extraction, called "Eigenface" [3]. In traditional PCA, the face images must be converted into 1D vector which has problem with high dimensional vector space.

Then, Yang et al. [12] has proposed an extension of PCA technique for face recognition using gray-level images. 2D-PCA treats image as a matrix and computes directly on the so-called image covariance matrix without image-to-vector transformation. The eigenvector estimates more accurate and computes the corresponding eigenvectors more efficiently than PCA. D. Zhang et al. [13] was proposed a method called Diagonal Principal Component Analysis (DiaPCA), which seeks the optimal projective vectors from diagonal face images and therefore the correlations between variations of rows and those of columns of images can be kept [3]. That is the reason why, in this paper, we used 2D-PCA (rows, columns and block-based) and DiaPCA (diagonal-based0 for extracting facial feature to be the input of Neural Network.

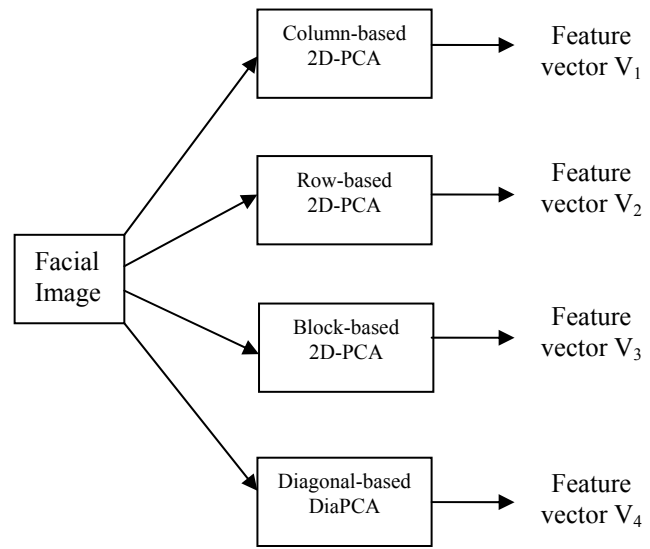

Fig 3. Facial Feature Extraction

Sub-Neural Network will classify the pattern based on the responsive feature. To compose the classified result, we can use the selection method, average combination method or build the reliability coefficients…

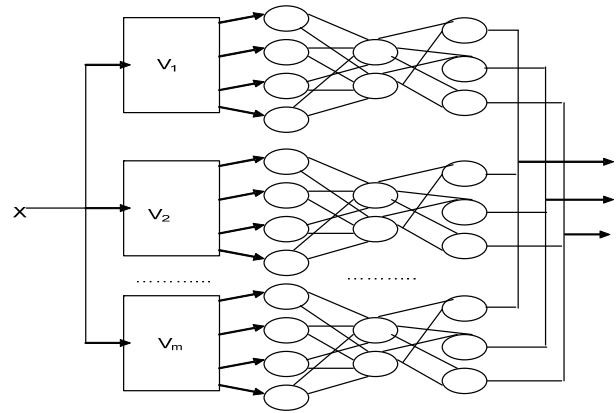

Fig 4. Processing of Sub Neural Networks

The selection method will choose only one of the classified results of a SNN to be the whole system's final conclusion:

$$P(\Omega i \mid X) = P_k(\Omega i \mid X) \quad (k=1..m) \qquad (1)$$

Where, $P_k(\Omega i \mid X)$ is the image X's classified result in the $\Omega i$ class based on a Sub Neural Network, $P(\Omega i \mid X)$ is the pattern X's final classified result in the Ωi. Clearly, this method is subjectivity and omitted information.

The average combination method [4] uses the average function for all the classified result of all SNN:

$$P(\Omega_i | X) = \sum_{k=1}^{m} \frac{1}{m} P_k(\Omega_i | X) \quad (2)$$

This method is not subjectivity but it set equal the importance of all image features.

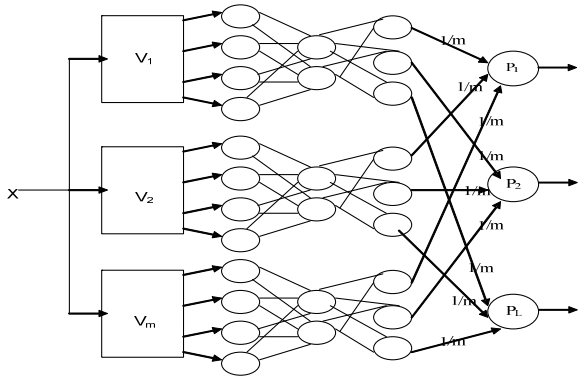

Fig 5. Average combination method

On the other approach is building the reliability coefficients attached on each SNN's output [4], [5]. We can use fuzzy logic, SVM, Hidden Markup Model (HMM) [6]… to build these coefficients:

$$P(\Omega_i | X) = \sum_{k=1}^{m} r_k P_k(\Omega_i | X) \quad (3)$$

Where, $r_k$ is the reliability coefficient of the $k^{th}$ Sub Neural Network. For example, the following model uses Genetics Algorithm to create these reliability coefficients.

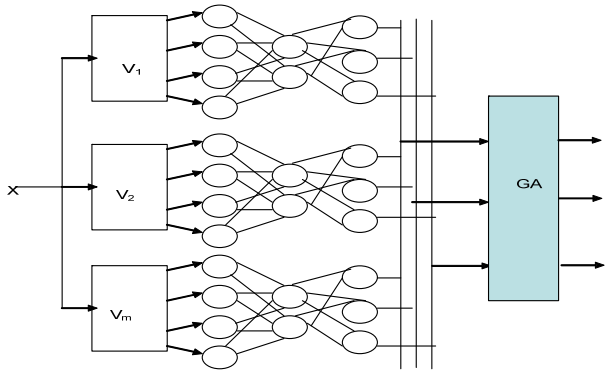

Fig 6. NN_GA model [4]

In this paper, we propose to use Neural Network technique. In details, we use a global frame consisting of some CNN(s). The weights of CNN(s) evaluate the importance of SNN(s) like the reliability coefficients. Our model combines many Neural Networks, called Multi Artificial Neural Network (MANN).

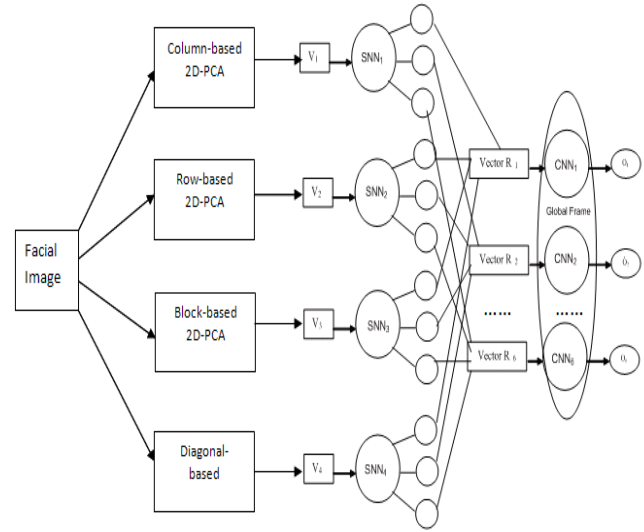

Fig 7. PCA and MANN combination

## 3. Image Feature Extraction using 2D-PCA

### 3.1 Two Dimensional Principal Component Analysis (2D-PCA)

Assume that the training data set consists of N face images with size of m x n. $X_1, X_2…, X_N$ are the matrices of sample images. The 2D-PCA proposed by Yang et al. [2] is as follows:

Step 1. Obtain the average image $\overline{X}$ of all training samples:

$$\overline{X} = \frac{1}{N} \sum_{i=1}^{N} X_i \quad (4)$$

Step 2. Estimate the image covariance matrix

$$C = \frac{1}{N} \sum_{i=1}^{N} (X_i - X)^T \times (X_i - X) \quad (5)$$

Step 3. Compute d orthonormal vectors $W_1, W_2, …, W_d$ corresponding to the d largest eigenvalues of C. $W_1, W_2,…, W_d$ construct a d-dimensional projection subspace, which are the d optimal projection axes.

Step 4. Project $X_1, X_2…, X_N$ on each vector $W_1, W_2, …, W_d$ to obtain the principal component vectors:

$$F_i^j = A_j W_i, \quad i=1..d; j=1..N \quad (6)$$

Step 5. The reconstructed image of a sample image Aj is defined as:

$$A_{recs(j)} = \sum_{i=1}^{d} F_i^j W_i^T \quad (7)$$

### 3.2 DiaPCA

The DiaPCA extract the diagonal feature which reflects variations between rows and columns. For each face image in training set, the corresponding diagonal image is defined as follows:

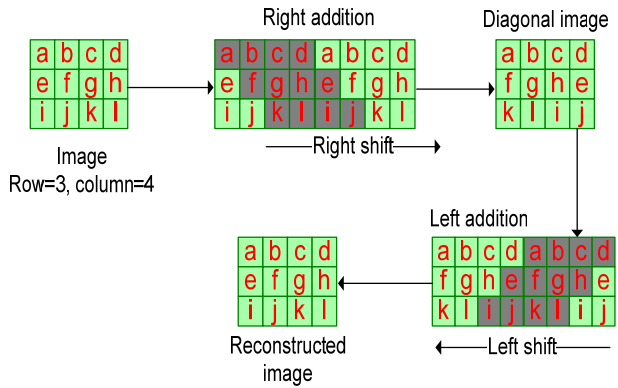

Fig 8. Extract the diagonal feature if rows ≤ columns

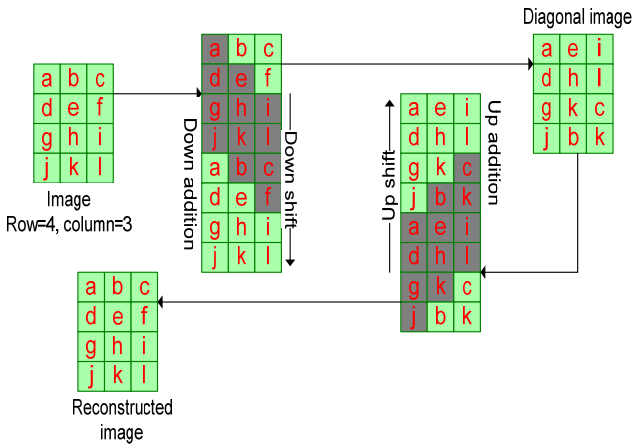

Fig 9. Extract the diagonal feature if rows > columns

### 3.3 Facial Feature Extraction

Facial feature extraction used 2D-PCA and its variants to project the pattern or image into different spaces with different grouping strategies. A facial image will be projected to 4 presentation spaces by PCA (column-based, row-based, diagonal-based, and block-based). Each of above presentation spaces extracts to the feature vectors.

So a facial image will be presented by $V_1$, $V_2$, $V_3$, $V_4$. In particular, $V_1$ is the feature vector of column-based image, $V_2$ is the feature vector of row-based image, $V_3$ is the feature vector of diagonal-based image and $V_4$ is the feature vector of block-based image.

Feature vectors ($V_1$, $V_2$, $V_3$, $V_4$) presents the difference orientation of original facial image. They are the input to Multi Artificial Neural Network (MANN), which generates the classified result.

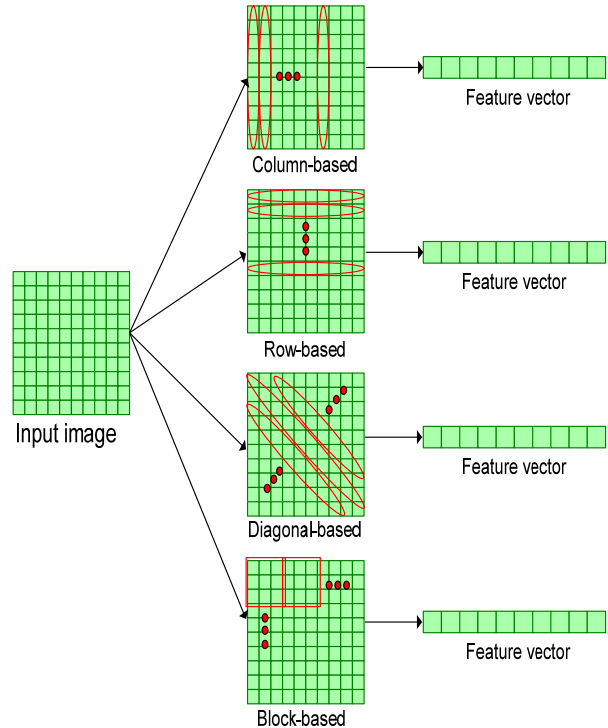

Fig 10. Facial Feature Extraction using 2D-PCA and DiaPCA

## 3. Multi Artificial Neural Network for Image Classification

### 3.1 The MANN structure

Multi Artificial Neural Network (MANN), applying for pattern or image classification with parameters (m, L), has m Sub-Neural Network (SNN) and a global frame (GF) consisting L Component Neural Network (CNN). In particular, m is the number of feature vectors of image and L is the number of classes.

*Definition 1*: SNN is a 3 layers (input, hidden, output) Neural Network. The number input nodes of SNN depend on the dimensions of feature vector. SNN has L (the number classes) output nodes. The number of hidden node is experimentally determined. There are m (the number of feature vectors) SNN(s) in MANN model. The input of the $i^{th}$ SNN, symbol is $SNN_i$, is the feature vector of an image. The output of $SNN_i$ is the classified result based on the $i^{th}$ feature vector of image.

***Definition 2***: Global frame is frame consisting L Component Neural Network which compose the output of SNN(s).

***Definition 3***: Collective vector $k^{th}$, symbol $R_k$ (k=1..L), is a vector joining the $k^{th}$ output of all SNN. The dimension of collective vector is m (the number of SNN).

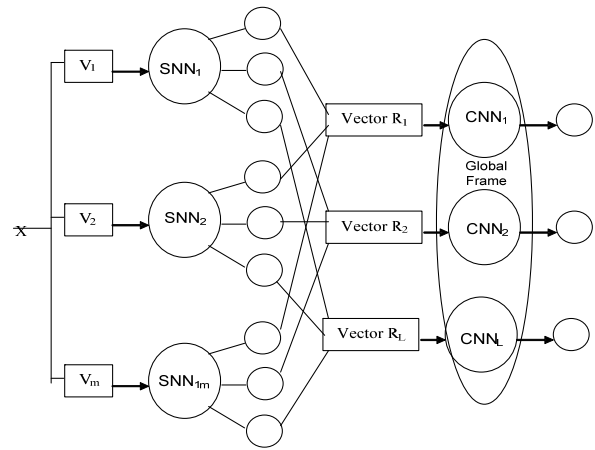

Fig 12. MANN with parameters (m, L)

### 3.2 The MANN training process

The training process of MANN is separated in two phases. Phase (1) is to train SNN(s) one-by-one called local training. Phase (2) is to train CNN(s) in GF one-by-one called global training.

In local training phase, we will train the $SNN_1$ first. After that we will train $SNN_2$, $SNN_m$.

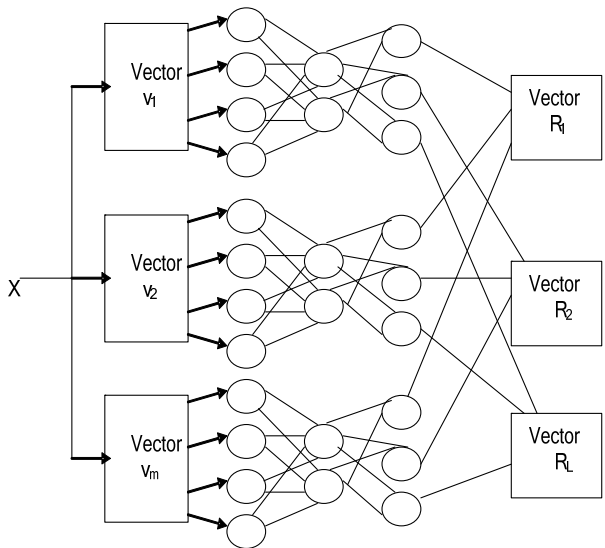

Fig 11. Create collective vector for CNN(s)

***Definition 4:*** CNN is a 3 layers (input, hidden, output) Neural Network. CNN has m (the number of dimensions of collective vector) input nodes, and 1 (the number classes) output nodes. The number of hidden node is experimentally determined. There are L CNN(s). The output of the $j^{th}$ CNN, symbols is $CNN_j$, give the probability of X in the $j^{th}$ class.

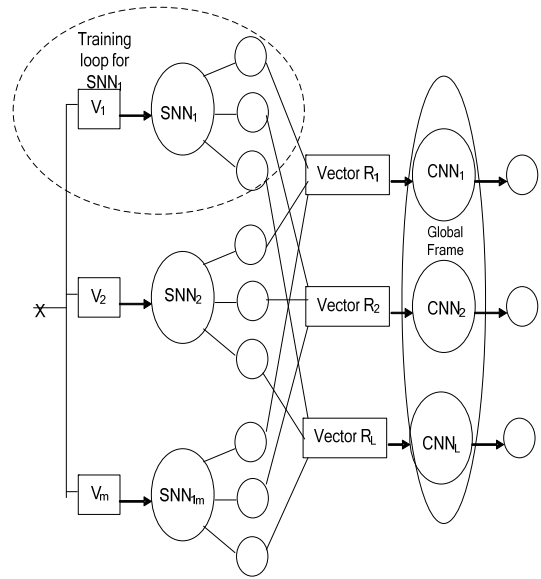

Fig 13. SNN1 local training

In the global training phase, we will train the $CNN_1$ first. After that we will train $CNN_2$,…, $CNN_L$.

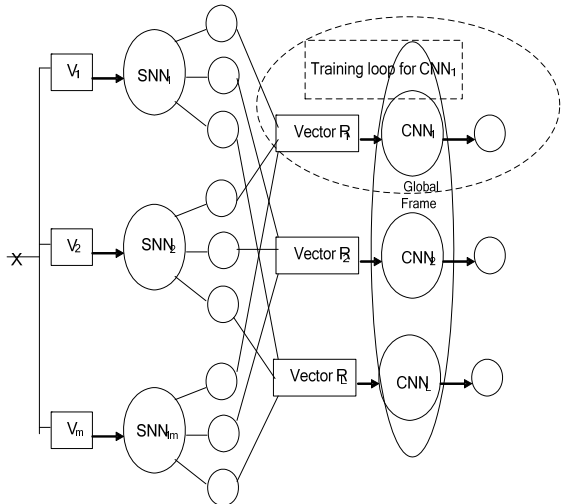

Fig 14. CNN$_1$ global training

### 3.3 The MANN classification

The classification process of pattern X using MANN is below: firstly, pattern X are extract to m feature vectors. The i$^{th}$ feature vector is the input of SNN$_i$ classifying pattern. Join all the k$^{th}$ output of all SNN to create the k$^{th}$ (k=1..L) collective vector, symbol R$_k$.

R$_k$ is the input of CNN$_k$. The output of CNN$_k$ is the k$^{th}$ output of MANN. It gives us the probability of X in the k$^{th}$ class. If the k$^{th}$ output is max in all output of MANN and bigger than the threshold. We conclude pattern X in the k$^{th}$ class.

## 4. Six Basic Facial Expressions Classification

In the above section, we explain the MANN in the general case with parameters (m, L) apply for image classification. Now we apply MANN model for six basic facial expression classifications. In fact that this is an experimental setup with MANN with (m=4, L=6).

We use an automatic facial feature extraction system using 2D-PCA (column-based, row-based and block based) and DiaPCA (diagonal-based).

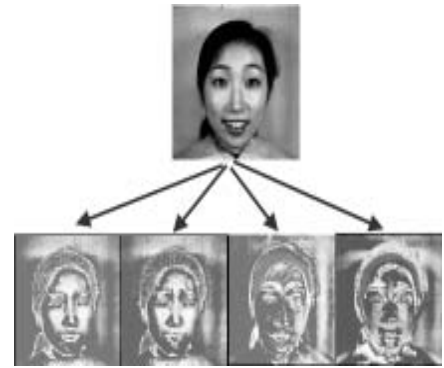

Fig 2. 2D-PCA and DiaPCA

The column-based feature vector is the input for SNN$_1$. The row-based feature vector is the input for SNN$_2$. The diagonal-based feature vector h is the input for SNN$_3$. The block-based feature vector is the input for SNN$_4$. All SNN(s) are 6 output nodes matching to 6 basic facial expression (happiness, sadness, surprise, anger, disgust, fear) [12]. Our MANN has 6 CNN(s). They give the probability of the face in six basic facial expressions. It is easy to see that to build MANN model only use Neural Network technology to develop our system.

We apply our proposal model for 6 basic facial expressions on JAFFE database consisting 213 images posed by 10 Japanese female models. The result of our experience sees below:

**Table 1. Facial Expression Classification Precision**

| Classification Methods | Precision of classification |
|---|---|
| SNN$_1$ | 81% |
| SNN$_2$ | 79% |
| SNN$_3$ | 86% |
| SNN$_4$ | 83% |
| Average | 89% |
| MANN | 93% |

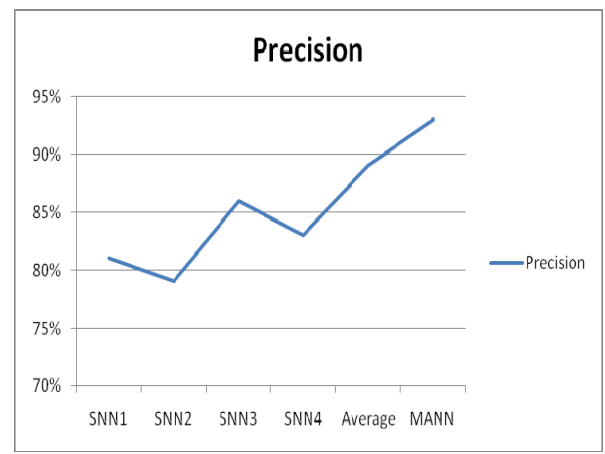

Fig 3. Facial Expression Classification Result

It is a small experimental to check MANN model and need to improve our experimental system. Although the result classification is not high, the improvement of combination result shows the MANN's feasibility such a new method combines.

We need to integrate with another facial feature sequences extraction system to increase the classification precision.

## 5. Conclusion

In this paper, we explain 2D-PCA and DiaPCA for facial feature extraction. These features are the input of our proposal model Multi Artificial Neural Network (MANN) with parameters (m, L). In particular, m is the number of images' feature vectors. L is the number of classes. MANN model has m Sub-Neural Network $SNN_i$ (i=1..m) and a Global Frame (GF) consisting L Components Neural Network $CNN_j$ (j=1..L).

Each of SNN uses to process the responsive feature vector. Each of CNN use to combine the responsive element of SNN's output vector. The weight coefficients in $CNN_j$ are as the reliability coefficients the SNN(s)' the jth output. It means that the importance of the ever feature vector is determined after the training process. On the other hand, it depends on the image database and the desired classification. This MANN model applies for image classification.

To experience the feasibility of MANN model, in this research, we propose the MANN model with parameters (m=4, L=3) apply for six basic facial expressions and test on JAFFE database. The experimental result shows that the proposed model improves the classified result compared with the selection and average combination method.

**Dr Le Hoang Thai** received B.S degree and M.S degree in Computer Science from Hanoi University of Technology, Vietnam, in 1995 and 1997. He received Ph.D. degree in Computer Science from Ho Chi Minh University of Sciences, Vietnam, in 2004. Since 1999, he has been a lecturer at Faculty of Information Technology, Ho Chi Minh University of Natural Sciences, Vietnam. His research interests include soft computing pattern recognition, image processing, biometric and computer vision. Dr. Le Hoang Thai is co-author over twenty five papers in international journals and international conferences.

**Tat Quang Phat** received B.S degree from Binh Duong University,

Vietnam, in 2007. He is currently pursuing the M.S degree in Computer Science Ho Chi Minh University of Science.
.
**Tran Son Hai** is a member of IACSIT and received B.S degree and M.S degree in Ho Chi Minh University of Natural Sciences, Vietnam in 2003 and 2007. From 2007-2010, he has been a lecturer at Faculty of Mathematics and Computer Science in University of Pedagogy, Ho Chi Minh city, Vietnam. Since 2010, he has been the dean of Information System department of Informatics Technology Faculty and a member of Science committee of Informatics Technology Faculty. His research interests include soft computing pattern recognition, and computer vision. Mr. Tran Son Hai is co-author of four papers in the international conferences and national conferences.